\documentclass[letterpaper, 10 pt, conference]{ieeeconf}  

\IEEEoverridecommandlockouts                              

\makeatletter
\let\NAT@parse\undefined
\makeatother
\usepackage[numbers,sort&compress]{natbib}

\usepackage[pdftex]{graphicx}
\usepackage{amsmath}
\usepackage{amssymb}  
\usepackage{multirow}
\usepackage{array,booktabs}
\usepackage{diagbox}
\usepackage{balance}
\usepackage{xspace}
\usepackage{algorithm}
\usepackage{algpseudocode}
\usepackage{adjustbox}
\usepackage{romannum}
\usepackage{subfloat}

 \usepackage[final]{hyperref}
 \hypersetup{
     colorlinks=true,
     linkcolor=blue,
     filecolor=magenta,
     urlcolor=blue,
     citecolor=black
 }
\usepackage[labelfont=bf, font=small]{caption}
\usepackage{./rpm_packages/rpm_SIunits}
\usepackage{./rpm_packages/rpm_acronyms}
\usepackage{./rpm_packages/rpm_math}
\usepackage{./rpm_packages/rpm_misc}

\usepackage{soul,color}
\usepackage{lipsum}

\usepackage{xcolor}

\usepackage{tikz} 
\usepackage{subcaption}
\usepackage{caption}
\title{\LARGE \bf Adaptive LiDAR-Radar Fusion for Outdoor Odometry \\Across Dense Smoke Conditions }     

\author{Chiyun Noh${}^{1}$ and Ayoung Kim${}^{1*}$
\thanks{$^\dagger$This paper was supported by Korea Institute for Advancement of Technology(KIAT) grant funded by the Korea Government(MOTIE) (P0020536, HRD Program for Industrial Innovation).}
\thanks{$^{1}$C. Noh and A. Kim are with the Dept. of Mechanical Engineering, SNU, Seoul, S. Korea {\tt\small [gch06208, ayoungk]@snu.ac.kr}}%
}
\begin{document}

\maketitle
\thispagestyle{empty}
\pagestyle{empty}

\begin{abstract}
Robust odometry estimation in perceptually degraded environments represents a key challenge in the field of robotics. In this paper, we propose a LiDAR-radar fusion method for robust odometry for adverse environment with LiDAR degeneracy. By comparing the LiDAR point cloud with the radar static point cloud obtained through preprocessing module, it is possible to identify instances of LiDAR degeneracy to overcome perceptual limits. We demonstrate the effectiveness of our method in challenging conditions such as dense smoke, showcasing its ability to reliably estimate odometry and identify/remove dynamic points prone to LiDAR degeneracy.

\end{abstract}
\section{Introduction \& Related Work}
\label{sec:intro}

\begin{figure}[!t]
    \centering
    \subfloat[LiDAR point cloud\label{fig:lidar}]{
        \includegraphics[width=0.5\columnwidth]{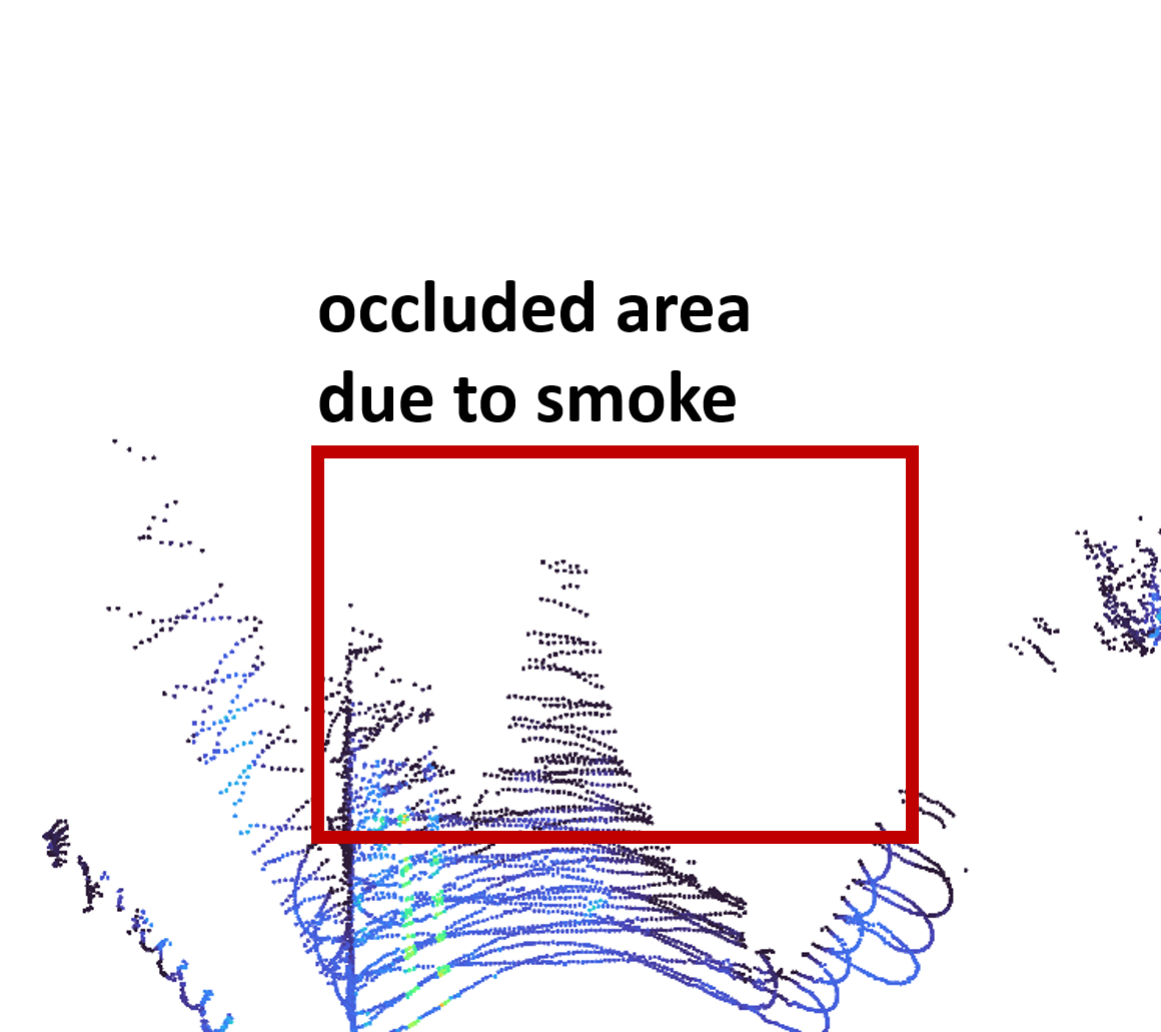}
    }
    \subfloat[Radar point cloud\label{fig:radar}]{
        \includegraphics[width=0.5\columnwidth]{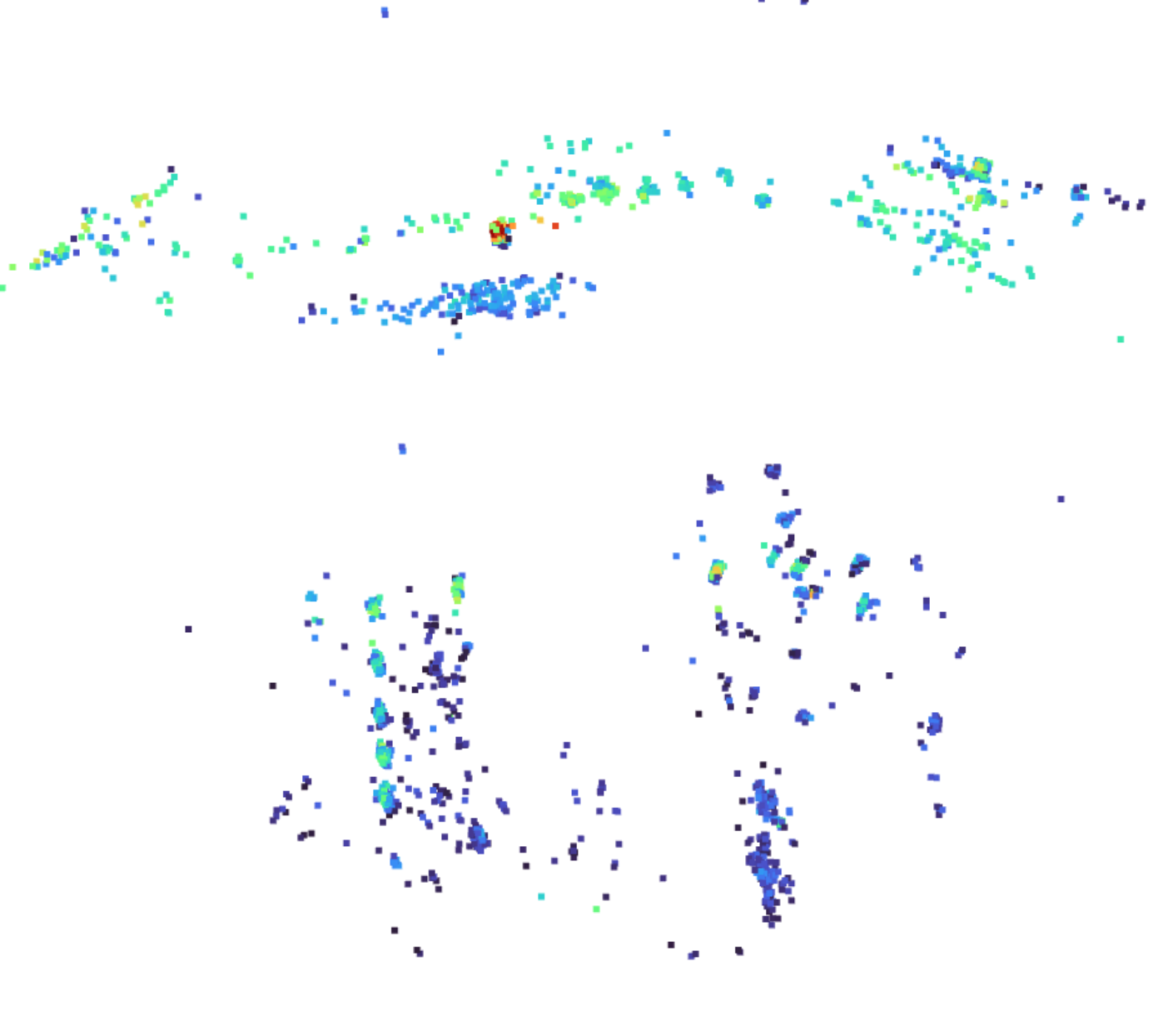}
    }
    \caption{Example of LiDAR and radar point clouds in smoke region. Occluded area is indicated by red boxes. As depicted in the figure, LiDAR point cloud exhibits occluded points due to dense smoke, whereas radar demonstrates robust perception. }
    \label{fig:fig1}
\end{figure}

Sensor fusion can enhance odometry estimation by leveraging complementary information from various sensors. In range sensing, LiDAR and radar effectively supplement each other's limitations. As shown in \figref{fig:fig1}, LiDAR cannot penetrate dense smoke or fog due to the relatively short wavelength of sensor, resulting in degraded perception performance \cite{two}, \cite{three}. On the other hand, radar shows promising robustness to such adverse conditions. As a result, there has been considerable amount of research conducted on odometry using 4D radar \cite{four}, \cite{five}, \cite{six}. 

Recent studies on radar odometry have shown amelioration from the use of Doppler velocity measurement in odometry estimation.  \citeauthor{four}~\cite{four} utilized estimated ego-velocity derived from ever-present static ground points, which demonstrated robust radar inertial odometry in high dynamic environments. 4D-iRIOM \cite{five} employed graduated non-convexity in conjunction with iterative extended Kalman filter for enhanced odometry estimation. \cite{six} used direct point cloud registration of radar scans using the Adaptive Probability Distribution GICP. Despite its robustness in harsh environments, extracting geometric features from radar point cloud is challenging due to the sparsity and noise of radar point clouds; this, in turn, leads to a degradation in robust odometry performance.

Recently, with the release of LiDAR/Radar Multi-Modal datasets focusing on degenerate environments, concurrent studies have begun leveraging sensor fusion under adverse conditions. \citeauthor{eight}~\cite{eight} studied a density-independent point cloud registration method between prior LiDAR maps and radar point clouds, which is specifically designed for low-visibility environments. Meanwhile, \cite{nine} suggested a tightly-coupled fusion approach using factor graphs to integrate LiDAR and radar data. These existing methods conducted experiments in indoor environments filled with smoke. In contrast, the NTU4DRadLM dataset \cite{twelve} used in this paper provides dense smoke environments, which limit the LiDAR perception. In dense smoke, inappropriate information such as smoke particles corrupts the LiDAR point cloud, potentially inducing divergence in odometry estimation.

The algorithm presented in this paper can detect area where LiDAR degeneracy occurs and enables robust odometry estimation by exclusively utilizing radar point cloud data.
By appropriately selecting radar point clouds through the three stages, robust odometry estimation is achieved in harsh environments. The contributions of this paper are as follows:

\begin{itemize}

    \item The algorithm proposed in this paper enables robust odometry estimation in dense foggy outdoor environments from LiDAR and radar fusion.
    \item Our proposed method can appropriately identify instances of LiDAR perception degeneracy and dynamic points in LiDAR point clouds. 
    \item Experimental validation using real-world data confirms the performance of the algorithm.

\end{itemize}

\section{Methodology}
\label{sec:relatedwork}

\begin{algorithm}[t!]
    \small
   \begin{algorithmic}[1]
    \caption{Sensor Select Algorithm}\label{euclid}
    \State \textbf{Input:} Time synced Radar Point Cloud $\mathcal{P}^R$, LiDAR Point Cloud $\mathcal{P}^L$ at time t with same reference frame.
    \State Split $\mathcal{P}^R$ into $\mathcal{P}^R_s$ and $\mathcal{P}^R_d$ using 3-Point RANSAC-LSQ \cite{ten}
    \For{ $\forall \mathbf{p}^j_{s} \in$ $\mathcal{P}^R_s$}
    \State find nearest point $\mathbf{p}^j_{L} \in \mathcal{P}^L$ from $\mathbf{p}^j_{s}$ using $\texttt{KDTree}$
    \If{$d(\mathbf{p}^j_{L}, \mathbf{p}^j_{s})<d_{th1}$}
    \State append $\mathbf{p}^j_{L}$ into $\mathcal{P}_{matched}$
    \EndIf
    \EndFor
    \State $\texttt{useLiDAR}$ = ($\frac{n(\mathcal{P}_{matched})}{n(\mathcal{P}^R_s)}>$ $r_{thres}$)
    \If{$\texttt{useLiDAR}$} 
        \State project $\mathcal{P}^L$, $\mathcal{P}^R_d$ into $xy$ plane and obtain $\mathcal{P}^L_{xy}$, $\mathcal{P}^R_{d, xy}$
        \For{ $\forall \mathbf{p}^j_{d, xy} \in$ $\mathcal{P}^R_{d, xy}$}
        \State find $\mathbf{p}^k_{L, xy} \in \mathcal{P}^L_{xy}$ within $R_{thres}$ from $\mathbf{p}^j_{d, xy}$
        \State append all pairs $(\mathbf{p}^k_{L, xy}, \mathbf{p}^j_{d, xy})$ into $\mathbf{P}^L_{select}$
        \EndFor
        \For{$\forall (\mathbf{p}^k_{L, xy}, \mathbf{p}^j_{d, xy}) \in \mathbf{P}^L_{select}$}
        \State calculate Covariance Matrix for $\mathbf{p}^j_{d, xy}$
        \State calculate $D_M(\mathbf{p}^k_{L, xy}, \mathbf{p}^j_{d, xy})$
        \If {$D_M < d_{th2}$}
        \State append $\mathbf{p}^k_{L}$ into $\mathcal{P}^L_{d}$
        \EndIf
        \EndFor        
        \State $\mathcal{P}^L_{s} = \mathcal{P}^L-\mathcal{P}^L_{d}, \;\mathcal{I} = \mathcal{P}^L_{s}$
    \Else
        \State $\mathcal{I} = \mathcal{P}^R_{s}$
    \EndIf 
    \State \textbf{return} $\mathcal{I}$
    \end{algorithmic}
\end{algorithm}

\begin{figure*}[t!]
    \centering
    \includegraphics[width=0.8\textwidth]
    {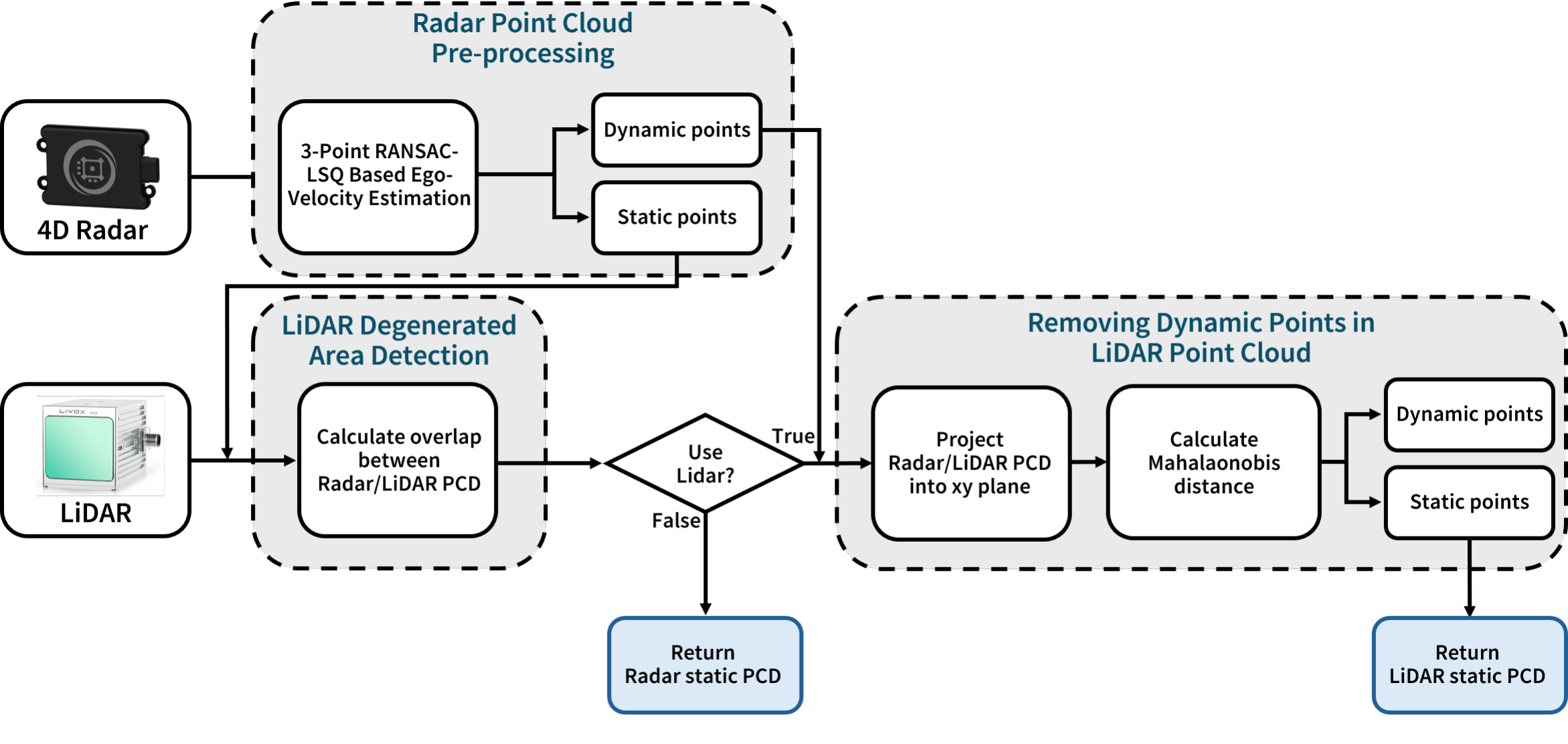}
    \caption{
    System diagram of proposed method. If the algorithm determines the LiDAR Degenerated Area, the radar static point cloud is used; otherwise, the LiDAR static point cloud is used as the input for the LIO method.
    }
    \label{fig:overview}
    \vspace{-3mm}
\end{figure*}

\subsection{Framework Overview}

The overall framework is depicted in \figref{fig:overview}. This paper proposes a new framework that utilizes radar as an alternative modality when LiDAR point cloud gets corrupted. The current methodology determines the input point cloud type for traditional LIO methods through three stages: 1) Radar Point Cloud Preprocessing 2) LiDAR Degenerated Area Detection 3) Removing Dynamic Points in LiDAR Point Cloud. 



\subsection{Radar Point Cloud Pre-processing}

4D radar measures both 3D position and Doppler velocity. Here, using Doppler velocity from radar allows us to infer ego velocity in which is used to distinguish dynamic point cloud and static point cloud in a single frame. In this paper, ego velocity is estimated using the 3-Point RANSAC-LSQ method introduced in \cite{ten}. Using estimated ego velocity, we can split radar point cloud $\mathcal{P}^R$ into static point cloud $\mathcal{P}^R_{s}$ and dynamic point cloud $\mathcal{P}^R_{d}$.
\begin{align}
    \begin{split}
 & \mathcal{P}^R = \mathcal{P}^R_{s}\cup \mathcal{P}^R_{d} \\[2pt]
& \mathcal{P}^R_{s} = \left\{\mathbf{p}^1_{s}, \; \mathbf{p}^2_{s}, \; \mathbf{p}^3_{s}, \; \cdots \;, \mathbf{p}^n_{s}\right\} \\[2pt]
& \mathcal{P}^R_{d} = \left\{\mathbf{p}^1_{d}, \; \mathbf{p}^2_{d}, \; \mathbf{p}^3_{d}, \; \cdots \;, \mathbf{p}^m_{d} \right\}
    \end{split}
\end{align}

\subsection{LiDAR Degenerated Area Detection} \label{sec:LiDAR_Disable_Area}
In environments with dense smoke or numerous dynamic objects, the difference between the actual static space and the LiDAR point clouds can induce failures in robust odometry estimation. On the contrary, leveraging the Doppler velocity from radar can sucessfully separate the point cloud into static and dynamic points. 

Comparing the static points from Radar point cloud $\mathcal{P}^R_{s}$ and the LiDAR point cloud $\mathcal{P}^L$ measures the degree of spatial difference within the LiDAR point cloud. For each radar point $\mathbf{p}^j_{s}$, we use nearest search algorithm to select the closest point $\mathbf{p}^j_{L}$ within the LiDAR point cloud. This process is repeated for all points in the $\mathcal{P}^R_{s}$ to obtain point set $\mathcal{P}^L_{set}$.
\begin{align}
   \mathcal{P}^L_{set} = \left\{\mathbf{p}^1_{L}, \; \mathbf{p}^2_{L}, \; \mathbf{p}^3_{L}, \; \cdots \;, \mathbf{p}^n_{L}\right\}
\end{align} 

The distances between points $\mathcal{P}^L_{set}$ and $\mathcal{P}^R_{s}$ are measured, and count the number of points below the threshold $d_{th}$.
\begin{align}
   \begin{split}
 & \mathcal{P}_{matched} = \left\{\mathbf{p}^k_{L}\; | \;\mathbf{p}^k_{L}\in \mathcal{P}^L_{set}, d(\mathbf{p}^k_{L}, \mathbf{p}^k_{s})<d_{th}\right\}  \\[2pt]
& n_{match} = n(\mathcal{P}_{matched})
    \end{split}
\end{align} 
where $d\left(\cdot \right )$ means Euclidean distance.

If the ratio of the $n_{match}$ to the total number of points in $\mathcal{P}^R_{s}$ exceeds a certain threshold, we consider the environment to be suitable for using LiDAR.

\subsection{Removing Dynamic Points in LiDAR Point Cloud}\label{sec:Removing_Points}

\begin{figure}[t!]
    \centering
    \includegraphics[width=0.5\columnwidth]
    {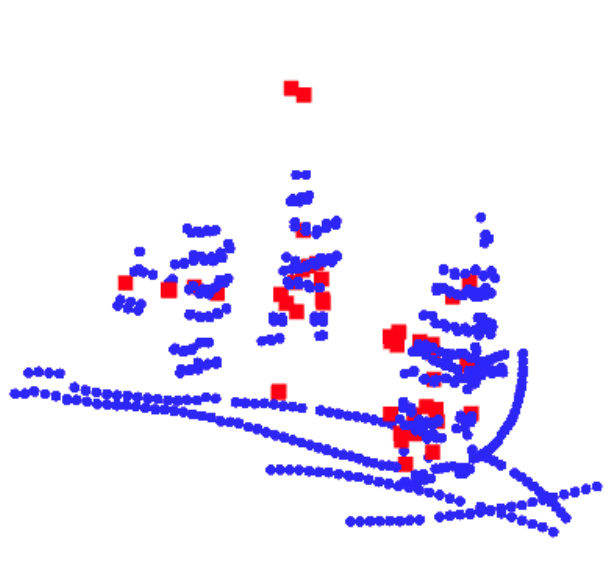}
    \caption{
    Example of radar dynamic point cloud $\mathcal{P}^R_{d}$ (red) and LiDAR point cloud $\mathcal{P}^L$ (blue). This scene depicts a scenario where a person is in motion.
    }
    \label{fig:fig3}
    \vspace{-0.3cm}
\end{figure}

Upon determining a suitable environment for using LiDAR, the next process is to remove the dynamic points from the LiDAR point cloud. Since LiDAR does not inherently contain temporal information, we have to use multiple LiDAR scans to remove dynamic points. However, by additionally using radar, dynamic points removal can be achieved in a single frame, as illustrated in \figref{fig:fig3}. 

Since achieving accurate correspondence between LiDAR and radar point clouds are infeasible due to their sparsity differences, we first project $\mathcal{P}^L$ and $\mathcal{P}^R_{d}$ onto the $xy$ plane and obtain $\mathcal{P}^L_{xy}$, $\mathcal{P}^R_{d, xy}$. Next, for each point in $\mathcal{P}^R_{d, xy}$, LiDAR points which is within a specific radius are selected, and this creates $\mathbf{P}^L_{select}$.

{\small
\begin{multline}
   \mathbf{P}^L_{select} = \{(\mathbf{p}^k_{L, xy}, \mathbf{p}^j_{d, xy})\; | \;\mathbf{p}^k_{L, xy}\in \mathcal{P}^L_{xy},\; \mathbf{p}^j_{d, ,xy}\in\mathcal{P}^R_{d, xy},\;\\
   d(\mathbf{p}^k_{L, xy}, \mathbf{p}^j_{d, xy})<R_{th}\} 
\end{multline}
}

According to \cite{six}, each point has an uncertainty with a standard deviation of $\sigma_r=0.00215r$, $\sigma_a=sin(0.5^{\circ})r$ and $\sigma_e=sin(1.0^{\circ})r$ for range, azimuth, and elevation, respectively. Through this, we can obtain the covariance matrix $S$ for each point, which is represented in the point's local frame. To transform this to the radar frame, we need to multiply the covariance matrix $S$ by the rotation matrix $R$, which allows us to obtain the covariance matrix $C$ for each point \cite{six}.
\begin{eqnarray}
\nonumber
C &=& RS\\
S &=& 
\begin{bmatrix}
\sigma_r & 0 & 0 \\
0 & \sigma_a & 0 \\
0 & 0 & \sigma_e \\
\end{bmatrix} \\ \nonumber
R &=& 
\begin{bmatrix}
\cos(\theta_a)\cos(\theta_e) & \sin(\theta_a) & -\cos(\theta_a)\sin(\theta_e) \\
-\sin(\theta_a)\cos(\theta_e) & \cos(\theta_a) & \sin(\theta_a)\sin(\theta_e) \\
\sin(\theta_e) & 0 & \cos(\theta_e) \\
\end{bmatrix}
\end{eqnarray}
where $\theta_a,\; \theta_e = \textit{point's azimuth} / \textit{elevation angle}$. 

However, we need to calculate the covariance within the 2D coordinate system. $xy$ plane's $2 \times 2$ covariance matrix can be derived from matrix $C$ by just eliminating both the third row and column \cite{eleven}. Through this, we can obtain the $2 \times 2$ covariance matrix for each point in radar, enabling us to measure the Mahalanobis distance between each LiDAR point and radar point $(\mathbf{p}^k_{L, xy}, \mathbf{p}^j_{d, xy})$ which is in $\mathbf{P}^L_{select}$. Base on the computed Mahalanobis distance, points $\mathbf{p}^k_{L, xy}$ within a specific distance are considered to be dynamic points which we then proceed to remove $\mathbf{p}^k_{L}$ in $\mathcal{P}^L$.
\begin{align}
    \begin{split}
   & \mathcal{P}^L_{d} = \{\mathbf{p}^k_{L} \;|\; (\mathbf{p}^k_{L, xy}, \mathbf{p}^j_{d, xy}) \in \mathbf{P}^L_{select},\\[2pt]
   & \;\;\;\;\;\;\;\;\;\;\;\;\;\;\;\;\;\;
   D_M(\mathbf{p}^k_{L, xy}, \mathbf{p}^j_{d, xy})<d_{th, M}\}\\[2pt]
   &\mathcal{P}^L_{s} = \mathcal{P}^L-\mathcal{P}^L_{d}
   \end{split}
\end{align}
where $D_M\left(\cdot \right )$ means Mahalanobis distance.

\section{Experiment and Result}
\label{sec:method}

\subsection{Trajectory Evaluation}
\subsubsection{Comparison Targets}\label{comparison}

we compared the proposed algorithm with the following methods:
\begin{itemize}
    \item \textbf{FAST-LIO2 \cite{seven}}: We estimated odometry solely from LiDAR-IMU data using one of the state-of-the-art LIO algorithms, FAST-LIO2. But, in areas where point clouds were not available due to smoke, we estimated the trajectory using only the IMU (\texttt{LiDAR}). 
    \item \textbf{4DRadarSLAM \cite{six}}: The most effective open-source 4D Radar Inertial odometry method currently known to us. (\texttt{Radar}).
\end{itemize}

\subsubsection{Dataset}
All experiments were evaluated using the NTU4DRadLM dataset \cite{twelve}. This dataset consists of six sequences, and we specifically utilized the $\texttt{cp}$ and $\texttt{garden}$ sequences, which are small-scale and low-speed. Although the dataset provides $\texttt{smoke}$ sequence, it does not include ground truth trajectories. Therefore, we simulated the effects of the dense smoke sequence by deleting LiDAR point cloud scans in certain sections of the $\texttt{cp}$ and $\texttt{garden}$ sequences, as shown in \figref{fig:fig4}.

\begin{table}[t]
\centering
\caption{RMSE result of APE and RPE. The smallest erros are highlighted in \textbf{Bold}. \texttt{LiDAR} method fails in every sequence, indicated by $-$. }
\label{tab:result}
\resizebox{0.95\columnwidth}{!}{
\begin{tabular}{c|c|c|c|c}
\toprule
\multicolumn{2}{c|}{} & {\parbox{2cm}{\centering\texttt{Ours}\\\texttt{(LiDAR+Radar)}}} & \texttt{Radar} & \texttt{LiDAR} \\ \midrule
\multirow{2}{*}{{\texttt{cp}}} & APE [m]& \textbf{1.133} & 2.142 & $-$ \\
& RPE [m]& \textbf{0.886} & 1.094 & $-$ \\
 \midrule
\multirow{2}{*}{{\texttt{garden}}} & APE [m]& \textbf{2.477} & 2.925 & $-$ \\
& RPE [m]& \textbf{0.737} & 0.817 & $-$ \\
\bottomrule
\end{tabular}}
\vspace{-2mm}
\end{table}

\subsubsection{Quantitative Result}
\begin{figure}[t!]
    \centering
    \subfloat[$\texttt{cp}$\label{fig:fig4_a}]{
        \includegraphics[trim={6mm 0 0 0},width=1\columnwidth]{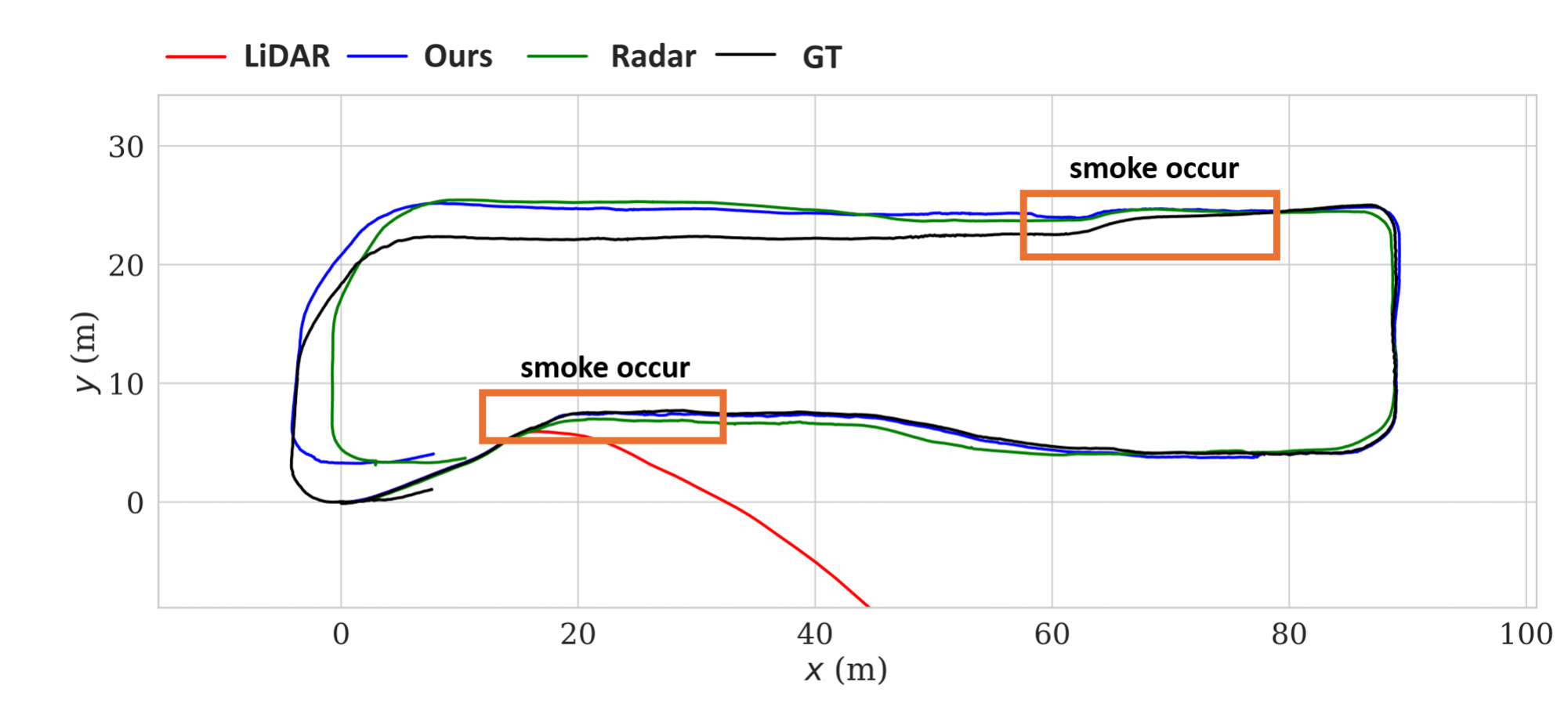}
    }\\%
    \vspace{1mm}
    \subfloat[$\texttt{garden}$\label{fig:fig4_b}]{
        \includegraphics[trim={6mm 0 0 0},width=0.95\columnwidth]{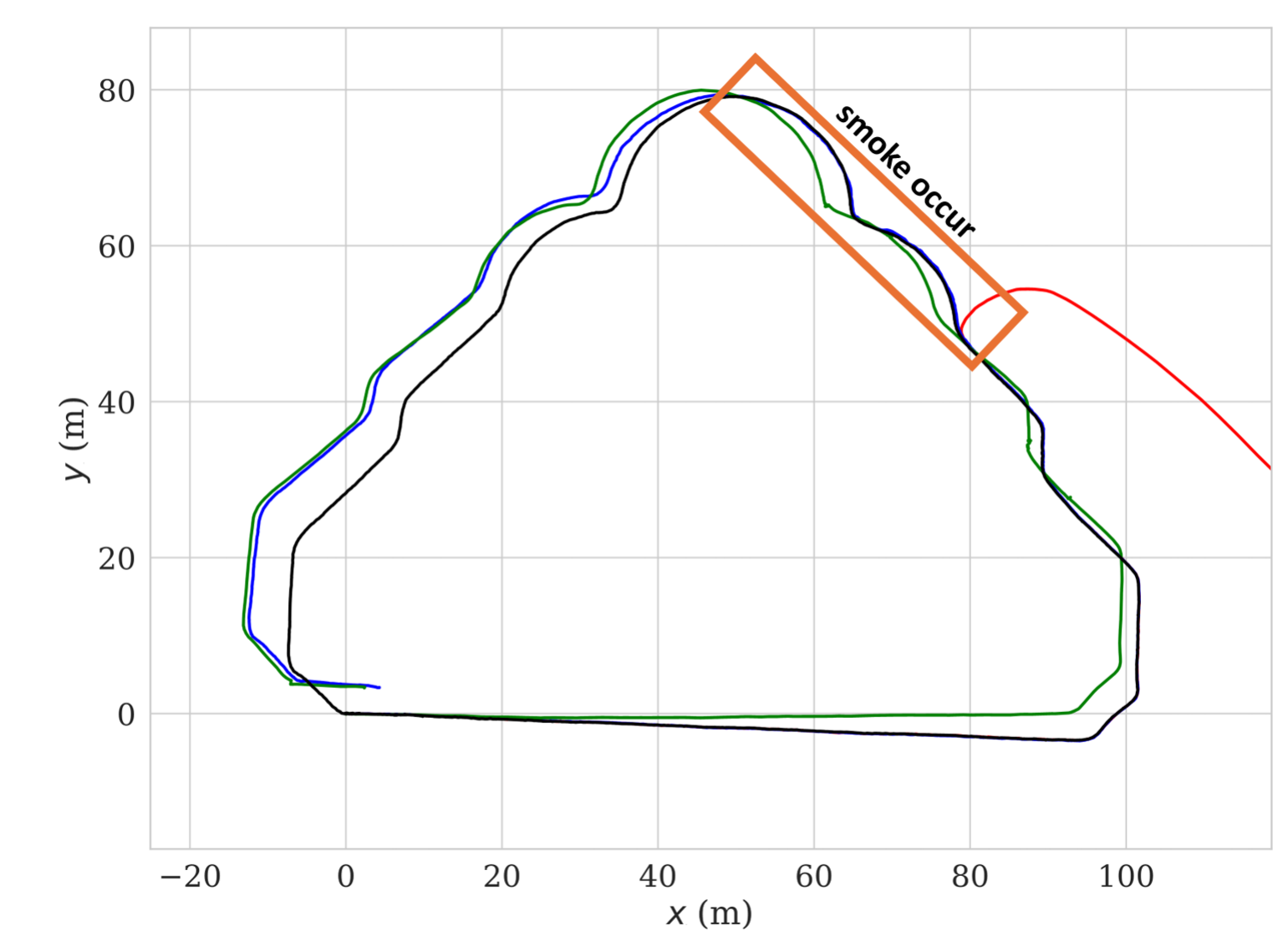}
    }\\
    \vspace{1mm}
    \subfloat[$\texttt{smoke}$\label{fig:fig4_c}]{
        \includegraphics[trim={6mm 0 0 0},width=1\columnwidth]{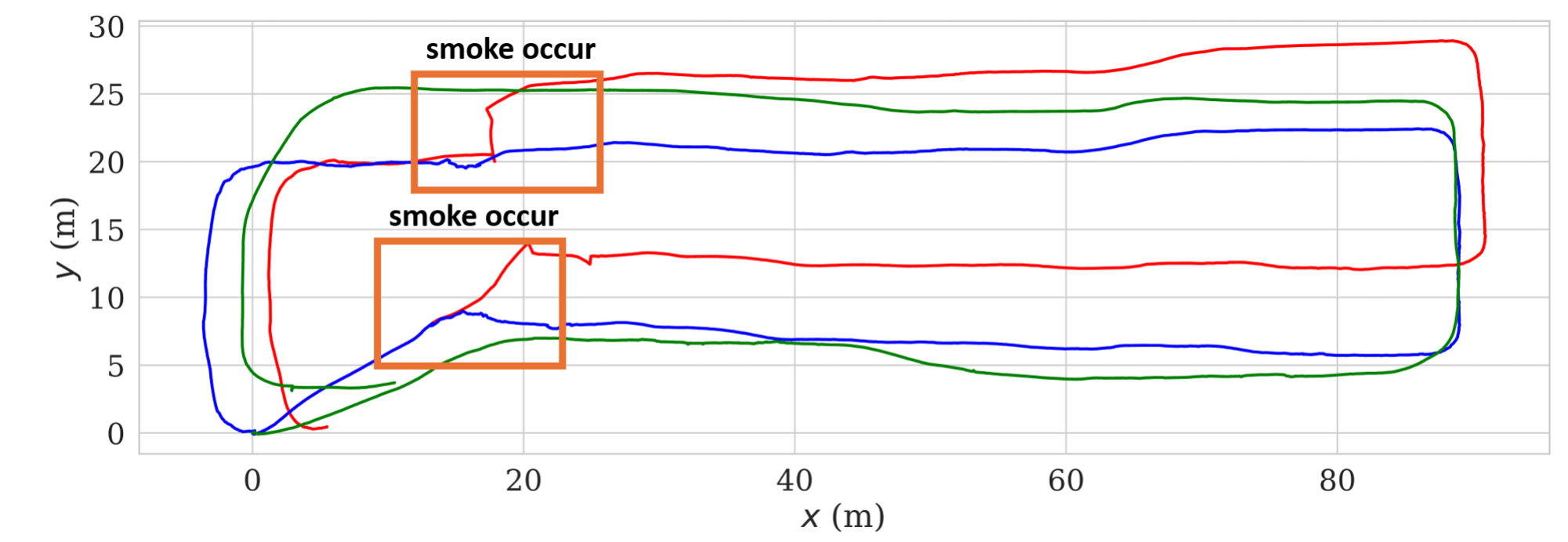}
    }
    \caption{
Trajectory result of sequence (a) $\texttt{cp}$, (b) $\texttt{garden}$ and (c) $\texttt{smoke}$. $\texttt{LiDAR}$, $\texttt{Ours}$, $\texttt{Radar}$ and $\texttt{GT}$ correspond to red, blue, green, and black, respectively. LiDAR scans were deleted within the Orange box area to simulate the smoke sequence effect. }
\label{fig:fig4}
\vspace{-2mm}
\end{figure}

Root Mean Square Error (RMSE) of Absolute Pose Error (APE) and Relative Pose Error (RPE) using \cite{thirteen} were used as evaluation metrics. The evaluation results are shown in \figref{fig:fig4} and \tabref{tab:result}. The $\texttt{LiDAR}$ method, which failed to perceive the environment due to smoke, did not demonstrate reliable performance and failed to predict the trajectory accurately. 

When comparing $\texttt{Radar}$ with the our proposed method, the proposed method exhibited higher odometry estimation performance in both sequences($\texttt{cp}$, $\texttt{garden}$). It is worth noting that while 4DRadarSLAM utilized Scan Context \cite{fourteen} for loop closure, the proposed algorithm did not conduct loop detection.

\begin{figure}[t!]
    \centering
    \subfloat[\label{fig:fig5_a}]{
        \includegraphics[trim={6mm 0 0 0},width=1\columnwidth]{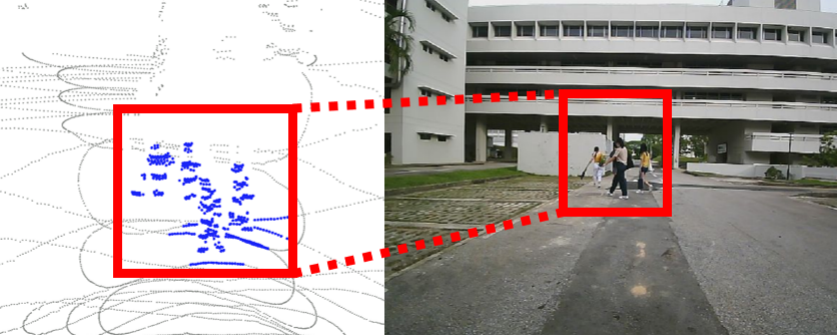}
    }\\%
    \vspace{2mm}
    \subfloat[\label{fig:fig5_b}]{
        \includegraphics[trim={6mm 0 0 0},width=1\columnwidth]{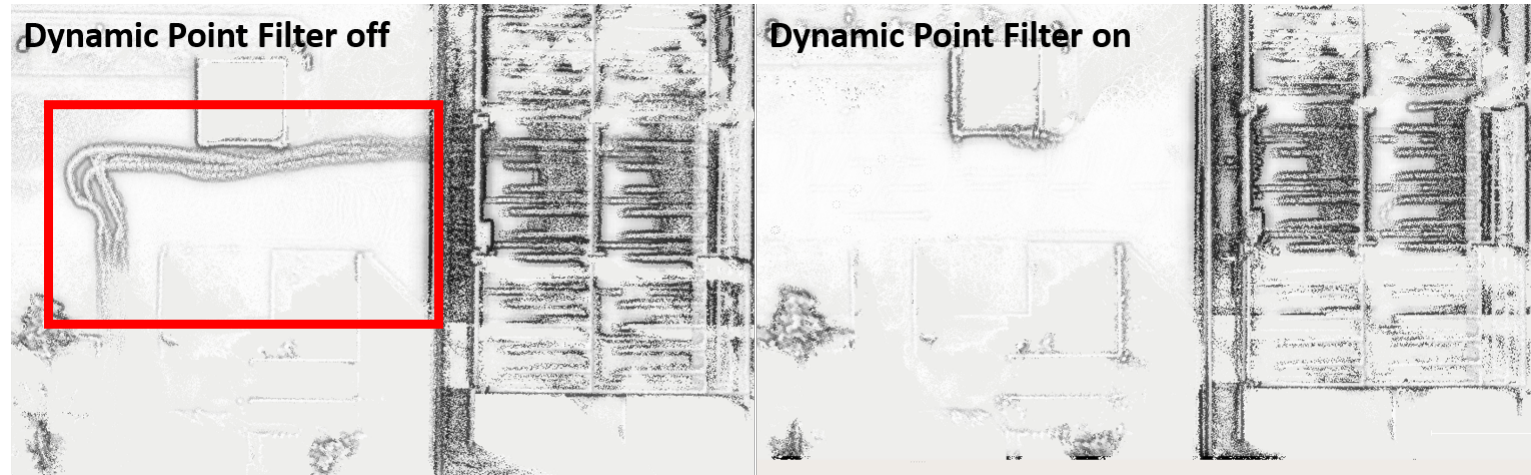}
    }
    \caption{Demonstration of the effect of $\textit{Removing}$ $\textit{Moving}$ $\textit{Points}$ $\textit{in LiDAR Point Cloud}$. In (a) left point cloud shows static LiDAR point cloud (gray) and dynamic LiDAR point cloud (blue) together. In (b), we use FAST-LIO2 to obtain map. $\mathcal{P}^L$ is used for input to make left map and $\mathcal{P}^L_{s}$ is used for input to make right map. After removing LiDAR point cloud, the trace (red box) is removed and map quality get better.
}
\label{fig:fig5}
\vspace{-2mm}
\end{figure}

\subsubsection{Qualitative Result}
Since ground truth poses are not available for the $\texttt{smoke}$ sequence, a quantitative comparison is not possible. However, when qualitatively evaluating the trajectory, as shown in \figref{fig:fig4_c}, method based solely on $\texttt{LiDAR}$ experiences severe drift in the dense smoke environment, whereas, the proposed algorithm estimates a robust trajectory.

\subsection{LiDAR Dynamic points Removal Evaluation}
The results of the dynamic points removal conducted in the section \ref{sec:Removing_Points} of this paper are depicted in \figref{fig:fig5_a}. As shown in \figref{fig:fig5_a}, by using removing module, moving people have been appropriately separated as dynamic points.

To assess mapping performance, we ran the   algorithm without the smoke region and obtained a map. The mapping results are depicted in \figref{fig:fig5_b}. When using $\mathcal{P}^L_{s}$ as a input, it can be confirmed that the trace created by people's movement has been deleted, and the quality of the static map has been improved by removing the dynamic point.

\subsection{LiDAR Degenerated Area Detection Evaluation}
We evaluated whether the algorithm appropriately detects perceptually degraded environment during timestamps in which the LiDAR point cloud is affected by smoke within the $\texttt{smoke}$ sequence. There were two main smoke time interval, and a total of 190 scans were affected by smoke within these sequences. Note that due to the flow of smoke, there are scans within the time sequence where LiDAR momentarily becomes possible for a while. Therefore, the exact number is not precise. Just for evaluation purposes, we assume that all scans between the beginning and end of the two time sequences are affected by the smoke. 

Through our algorithm, perceptual degraded environment were detected in 150 scans through our algorithm, resulting in a recall of 0.79. Despite such limitations, the algorithm robustly detects LiDAR degenerated areas.

\section{Conculsion \& Future Work}
\label{sec:conculsion}

This paper presents an adaptive Lidar-radar sensor fusion method for robust odometry performance in dense foggy outdoor environments. From quantitative and qualitative evaluation on NTU4DRadLM dataset, our proposed method demonstrated reliable odometry estimation in challenging conditions. In addition, our method consistently identifies regions where LiDAR degeneracy occurs while eliminating dynamic points in LiDAR point cloud successfully. 

The method proposed in this paper exhibits limitations in accurate scan registration due to the disparities in sensor modalities between LiDAR and radar. Future work will be focused on robust scan registration estimation between two different modalities and conducting experiments in diverse environments to further enhance algorithm performance.


\balance
\small
\bibliographystyle{IEEEtranN} 
\bibliography{string-short,references}

\end{document}